# IMAGE ORDINAL CLASSIFICATION AND UNDERSTANDING: GRID DROPOUT WITH MASKING LABEL

*Chao Zhang*[1,2], *Ce Zhu*[1], *Jimin Xiao*[3], *Xun Xu*[4*], *Yipeng Liu*[1]

[1]University of Electronic Science and Technology of China
[2]Sichuan Police College
[3]Xi'an Jiaotong-Liverpool University
[4]National University of Singapore
galoiszhang@gmail.com, {eczhu, yipengliu}@uestc.edu.cn,
jimin.xiao@xjtlu.edu.cn, elexuxu@nus.edu.sg

## ABSTRACT

Image ordinal classification refers to predicting a discrete target value which carries ordering correlation among image categories. The limited size of labeled ordinal data renders modern deep learning approaches easy to overfit. To tackle this issue, neuron dropout and data augmentation were proposed which, however, still suffer from over-parameterization and breaking spatial structure, respectively. To address the issues, we first propose a grid dropout method that randomly dropout/blackout some areas of the training image. Then we combine the objective of predicting the blackout patches with classification to take advantage of the spatial information. Finally we demonstrate the effectiveness of both approaches by visualizing the Class Activation Map (CAM) and discover that grid dropout is more aware of the whole facial areas and more robust than neuron dropout for small training dataset. Experiments are conducted on a challenging age estimation dataset – Adience dataset with very competitive results compared with state-of-the-art methods.

## 1. INTRODUCTION

The task of image ordinal classification is a special case of generic classification problem, which takes input data labeled with ordinal scalar value. The scalar value is often discrete but carries ordering information. For example, the task of classifying a dog and a cat needs to separate two different categories. However ordinal classification, such as age estimation, needs to separate faces of different ages, i.e., all the objects are faces. Such problem is often cast as a classification

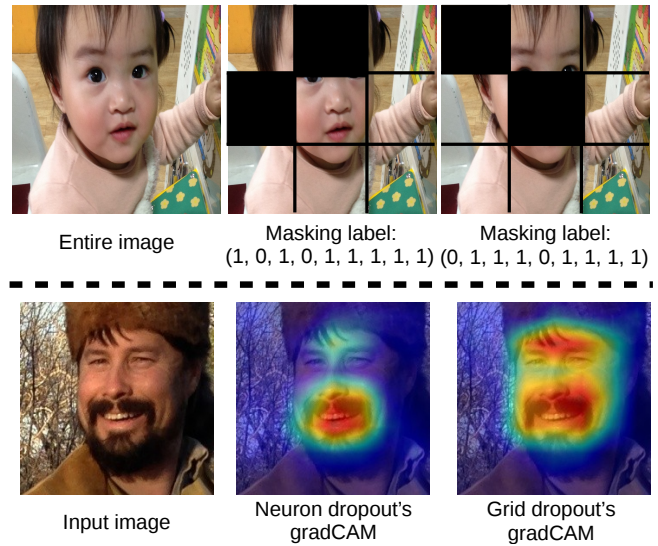

**Fig. 1**. Above: image ordinal classification with randomly blackout patches. It is easy for human to recognize the age regardless of the missing patches. The masking label is also useful to image classification. Bottom: grid dropout's gradCAM is better than that of neuron dropout. That is to say, grid dropout can help learning feature representation.

problem [1]. With the proliferation of convolutional neural network (CNN), works have been carried out on ordinal classification with CNN [1][2][3]. Though good performances have been logged with modern deep learning approaches, there are two problems in image ordinal classification. On one hand, the amount of ordinal training data is very limited which prohibits training complex models properly, and to make matters worse, collecting large training dataset with ordinal label is difficult, even harder than labelling generic dataset. Therefore, insufficient training data increases the risk of overfitting. On the other hand, less studies are conducted to understand what deep models have learned on ordinal data

* corresponding author

This research is supported by National Natural Science Foundation of China (NSFC, No. 61571102, No. 61602091, No. 61372187), Applied Basic Research Programs of Science and Technology in Sichuan(No. 2018JY0035).

than other generic classification tasks. Nowadays CNN models are no longer a black box [4][5][6]. We can understand their mechanism by visualizing and analyzing neurons' activation. However, this is not properly studied in the ordinal data classification.

To tackle the CNN overfitting issue, dropout was proposed as an efficient solution [7], which randomly drop units from the neural network during training. We therefore refer it to neural dropout throughout this work. Neuron dropout can be viewed as a data augmentation method. From another perspective, it is also regarded as an ensemble method that includes many sub-networks. Without neuron dropout, CNN may seriously suffer from the overfitting problem as the number of training iteration increases. Alternative to neuron dropout, some existing works implement data augmentation [8] to expand the size of training data for ordinal data classification. For example, randomly rotating, translating and cropping images into many patches are popular data augmentation methods with good performance [8][9]. However, it may break image's spatial structure, and lose global information. Proper data augmentation method is important for generalization ability.

Visualizing certain neuron activations has been studied for revealing the spatial localization of object. Recently, class activation map (CAM) [4] and gradient-weighted class activation map (gradCAM) [5] both propose to understand and visualize class specific information in image classification. In generic classification, CAM/gradCAM is able to highlight the most discriminative areas for each class. Thanks to the gradCAM visualization, we further identified that neuron dropout is very likely to overfit to certain parts of the face, e.g. forehead, mouth or nose, as illustrated in Fig. 1. However, we generally agree that a single area of interest can hardly discriminate age which is often correlated with many facial features.

Based on above observations, we propose a grid dropout method that can simultaneously improve the ordinal classification accuracy by reducing overfitting and obtain better finegrained class activation maps for visualization. Our model randomly overlays a masking label (as shown in Fig. 1) for training images, thus expanding the training dataset to avoid overfitting. More importantly, it does not break spatial structure of image. Moreover, randomly dropping out certain patches of the image can reduce the risk of over-weighting specific areas of image because we believe it is necessary to take the whole discriminative areas into consideration rather than focusing on certain small parts. The effect grid dropout is further supported by the visualization of gradCAM. Lastly, we incorporate an additional objective/loss, i.e. predicting the masking labels, into our model. We believe that if the model is able to learn discriminative features from facial parts, it should be able to tell the blackout patches. This ability should in turn benefit the main task of ordinal classification. Our primary contributions are as follows:

- We propose a grid dropout method that drops some of the image patches. It reduces the overfitting problem, and leads to better ordinal classification accuracy.
- We propose to use a masking label to provide supervision on the spatial information in the training stage. It turns out that the use of masking label brings extra performance gains for ordinal classification.
- gradCAM is adopted to visualize the discriminating areas of images and it suggests grid dropout is better than neuron dropout in capturing the global facial features.
- We finally analyze the relationship between neuron dropout and grid dropout, and reveal that grid dropout is more suitable for small or medium dataset like age estimation, while neuron dropout needs large dataset for training.

## 2. RELATED WORK

### 2.1. Ordinal Classification

Ordinal classification takes input data with scalar value labels that are often discrete but carry the ordering information between categories. Many applications such as age estimation [3][10][2] and facial beautiness [9] fall into the task of ordinal classification. Image ordinal estimation generally uses hand-crafted features and individual classifier or regressor in traditional methods. With the increasing popularity of deep learning, many recent works apply CNNs [11][1] and achieve better performance. However, deep learning models usually need large amount of training data. Collecting sufficient training data with ordinal label is hard compared with generic classification problems. There are three directions towards this issue. Firstly, some previous works [8] use patch-based method to improve recognition accuracy and reduce overfitting. Though this can greatly augment the dataset, it breaks the image's spatial structure only with some local patches. Secondly, some works are based on multiscale deep model [8] and different generic models such as residual model [12]. Finally, others apply different loss functions, such as triplet loss, weighted loss, and multi-task loss and so on.

### 2.2. Image visualization for Image Classification

In order to understand why CNNs perform so well, many works [6][13][4][5] are proposed to explain generic image classification. Toward generic understanding, [6] first introduces a novel visualization technique that gives insight into the function of intermediate feature layers and the operation of the classifier. [4] proposes CAM that localizes the specific object in image, which allows explanations for a specific class of image classification. [5] comes up with an efficient generalization of CAM, known as gradCAM, which fuses class discriminative property of CAM with existing pixel-space gradient visualization techniques such as guided back-propagation. However, there are relatively few works dealing with deep ordinal classification, let alone the analysis into understanding the spatial neuron activation.

## 3. PROPOSED MODELS

### 3.1. Grid Dropout

In this work, we propose to exploit randomly generated mask to blackout certain areas of the image for the purpose of dropout and data augmentation during the training phase. Firstly, we partition the image into $s \times s$ equal size grids. Then, a certain proportion of grids are randomly dropped out. In this way, grid dropout continuously produces multiple different samples for one image. It enlarges the training dataset as a data augmentation method. Continuously inputting different samples of the same image into a deep model can greatly reduce overfitting. More importantly, unlike the random cropping it also maintains the spatial information for the training image.

### 3.2. Grid Dropout with Masking Label

CNNs had been seen as a black box for many tasks at early ages, i.e., inputting image and outputting classification probabilities. Without clear understanding of the mechanism, naive data augmentation usually loses the spatial information. In this paper, a method based on masking label is proposed to preserve such spatial information. In the training process, each image is fed into the deep model many times with different blackout (dropped) masks. When the image with random mask is fed into the model, locations of the brighten cells (non-dropped) are known. We fully exploit such information to learn the image's spatial organization (as shown in Fig. 1).

The masking label records the the locations of the selected and dropped grids. For example, in the middle image of Fig. 1, the second and fourth grids are dropped (the grid number is counted row-wise). The masking label for this image is $(1, 0, 1, 0, 1, 1, 1, 1, 1)$. With both category label and masking label, the deep model can be written as $\mathcal{F} : \mathcal{I} \to (\mathcal{Y}, \mathcal{H})$ given a set of training images and labels $\{(I_n, y_n, h_n), n = 1, 2, \cdots, N\}$, where $y_n$ and $h_n$ represent image category label and masking label. We construct the total loss as,

$$L = L_{cla}(X_n, y_n) + \beta L_{mask}(X_n, h_n) \quad (1)$$

where $X_n$, $L_{cla}$ and $L_{mask}$ are feature vector, loss functions for ordinal classification and masking classification, respectively. For $L_{cla}$ and $L_{mask}$, we choose softmax loss function and cross entropy sigmoid loss function, respectively. $\beta$ is a hyper-parameter to balance the two losses, and it is set as 0.5 by default.

In addition, ordinal classification can consider both the class category and labels' ordinal relationship. The training objective is therefore composed of a weighted sum of three losses:

$$L = L_{cla}(X_n, y_n) + \alpha L_{reg}(X_n, y_n) + \beta L_{mask}(X_n, h_n) \quad (2)$$

where $L_{cla}$, $L_{reg}$, $L_{mask}$ represent the softmax loss, Euclidean loss and cross entropy loss, respectively, $\alpha$ and $\beta$ are both set to 0.5 in the subsequent experiment.

### 3.3. Problems of Neuron Dropout for Ordinal Classification

Neuron drop-out was proposed in [7] by randomly dropping out hidden neurons in each training iteration. The drop-out process resembles a boosting or ensemble technique. Each training iteration can be seen as a weak classifier, and the whole training process leads to an ensemble model. With sufficient images, ensemble model can potentially improve performance. However, ensemble classifier boosts large number of parameters to learn. For example, a neuron dropout layer with $D$ units can be seen as a collection of $2^D$ possible sub-networks/weak classifier. In contrast, grid dropout includes only one classifier while dealing with diversified inputs. Therefore, less data is required for training good models by grid dropout in comparison with neuron dropout.

From the perspective of CNN visualization, the defect of neuron dropout is revealed by CAM/gradCAM that comes from rigorous mathematical derivation. Let $F_k$ represents the $k$-th channel feature map in the last convolutional layer with $l \times l$ size, and $\mathbf{W} = \{w_{kc}\}_{K \times C}$ denotes the weight matrix of the classification layer, where $C$ is the number of classes, $K$ is last feature map channel number. Essentially, $w_{kc}$ indicates the importance of $F_k$ for class $c$, and can be calculated by the gradient of $y^c$ with respect to feature maps $F_k$,

$$w_{kc} = \sum_{(m,n)} \frac{\partial y^c}{\partial F_k(m, n)} \quad (3)$$

The activation map for each class $S_c$ is denoted as $S_c = \sum_k w_{kc} F_k$, where $S_c$ is of $l \times l$ size. In neuron dropout, half neurons are randomly dropped out, i.e. half of $w_{kc}$ are set to zero. Therefore, the equation (Eq. (3)) no longer holds true for neuron dropout.

### 3.4. Analysis on Generalization

The best way to enhance deep model's generalization ability is to train it on large dataset. A complex model usually obtains perfect performance on the training images, however, it is possible to perform bad on the testing images. Our proposed grid dropout aims to improve model's generalization by enlarging dataset size.

Generalization error, here representing testing error, is an important index that evaluates generalization ability. It is measured by two meaningful components: bias and variance. In general, bias measures the training loss and variance describe model's complexity. When the model becomes complex, its variance becomes large. As a common sense, bias-variance trade-off between accuracy and stability is always there in deep learning. Using the grid dropout, each training image has $C_{s \times s}^{[s \times s \times p]}$ choices, where $s$ is the number of partition on row and column and $p \in [0, 1]$ is the ratio of selected grids. Grid dropout is capable of enlarging existing dataset, and slowing down the variance's growth (as shown in Fig. 2 and the experiment results in Fig. 4). From Fig. 2, the slow growth of the generalization error is preferred for ideal training.

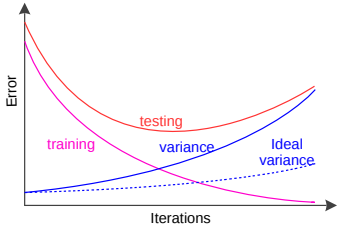

**Fig. 2**. In supervised learning, we always want to achieve optimal status with the blue dashed curve. However, in most cases, the realizable ones are with blue solid curve, especially for the condition of insufficient data. (Best viewed in color.)

## 4. EXPERIMENTS

In this section, we compare the ordinal group classification performance for neuron dropout [7], grid dropout and grid dropout with masking label. We also compare gradCAM using neuron dropout [7] and grid dropout. We use the challenging Adience dataset [1], which partitions the age interval into 8 levels. Many previous works [1][2][3] are based on this dataset, while few of them are under the same conditions and with the same network model. Some works use pretrained model with facial dataset such as MS-Celeb-1M [3].

### 4.1. Dateset

The work [1] uses Adience to perform group classification. Adience gives different levels $y_n$, in which person ages are divided into eight levels $\{0, 1, \cdots, 7\}$ (0-2, 4-6, 8-13, 15-20, 25-32, 38-43, 48-53, 60-). Adience includes roughly 26,000 images from 2,284 persons. Images in Adience are unconstrained, which makes the experiment very challenging. We follow the standard protocol [1] to perform a 5-fold cross validation, which are denoted as Cross0, Cross1, Cross2, Cross3, Cross4. We evaluate different models' ability by general recognition accuracy, generalization ability, visualization of gradCAM.

### 4.2. Experimental Settings

The work [1] defines their own specific network structure for ordinal group classification. In order to easily reproduce our results and to be fair, we use VGG-net [14]. We conduct all the experiments under the same conditions.

In the model training step, we initialize the learning rate with 0.001. We apply an exponential decay function, i.e., decay every 5000 steps with a base of 0.5. The batch size is 64, and we train all the models 150 epochs. In the neuron dropout experiment, the dropout rate is fixed as 0.5. For all models, we freeze some beginning layers from `Conv1-1` to `Conv2-2`. Grid dropout partitions each image into $5 \times 5$ grids with the same size. The proportion of the dropped grids is 0.25, which diversifies Adience $C_{5 \times 5}^{[5 \times 5 \times 0.25]}$ times.

For ordinal group classification performance comparison, we conduct three groups of experiments. The first group of experiments compare neuron dropout, neuron+grid dropout, and neuron+grid dropout with masking label. In this group, VGG-net use pretrained parameters from ImageNet. The second set of experiments compare neuron dropout and grid dropout without fine-tuning for the fully connected layers. Since we want to understand which model has strong generalization ability and better performance, training fully connected layers from scratch is a good way. The last set of experiment compares our method with the state-of-the-art works even though our methods can be extended to these works.

### 4.3. Neuron Dropout, Grid Dropout and Masking Label

In order to show grid dropout's efficacy, we compare 3 scenarios, including neuron dropout, neuron+grid dropout, and neuron+grid dropout with masking label. We show the model's ability from three aspects: classification accuracy, training and testing loss curves, and visualization of gradCAM.

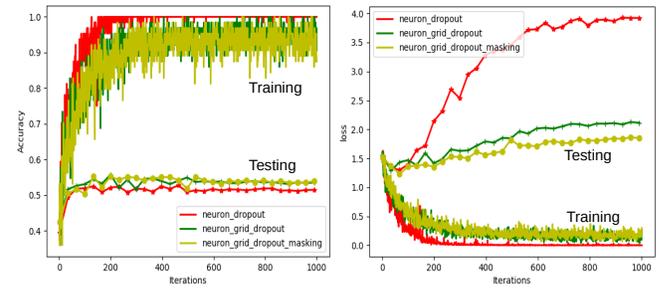

**Fig. 3**. Comparison of accuracy and loss curves for the three methods. (Best viewed in color and magnifier.)

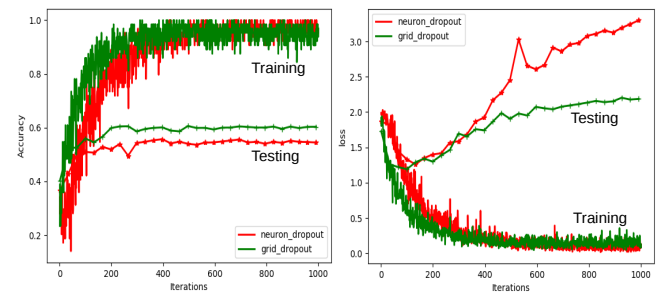

**Fig. 4**. Comparison of accuracy and loss curves for neuron dropout and grid dropout. (Best viewed in color and magnifier.)

In Table 1, the results show that neuron+grid dropout has higher accuracy than that of neuron dropout, and neuron+grid dropout with masking label has higher accuracy than that of neuron+grid dropout. This demonstrates that grid dropout and masking label both work well and effectively reduce overfitting. Without loss of generality, we randomly choose a split with training process shown in Fig. 3. Compared with neuron dropout, the gap between the training loss and testing loss in neuron+grid dropout and neuron+grid+masking present more close trend with the ideal variance curve described in Fig. 2. From both the accuracy and loss curve, neuron dropout falls into overfitting very early, and becomes very serious as the number of iterations increases[1]. In contrast, grid dropout with masking label performs the best out of all alternatives with simple grid dropout already performing very competitively. In addition, we can see in Fig. 5 the activation map (gradCAM) for neuron+grid dropout with mask label (third row) get more focused on the entire face area than without grid dropout (second row). This suggests the grid dropout with masking label is able to discover all discriminative areas for ordinal classifications.

**Table 1**. The results of neuron dropout, grid dropout and grid dropout with masking label.

| models | Cross0 | Cross1 | Cross2 | Cross3 | Cross4 | Mean |
|---|---|---|---|---|---|---|
| neuron dropout (%) | 61.17 | 41.66 | 57.83 | 49.68 | 51.45 | 52.36 |
| neuron+grid dropout (%) | 62.44 | 44.31 | 57.80 | 49.08 | 53.51 | 53.43 |
| neuron+grid+masking (%) | 63.14 | 45.79 | 57.53 | 49.17 | 53.94 | 53.91 |

**Table 2**. Comparative results of neuron dropout and grid dropout. Fully connected layers are trained from scratch without fine-tuning.

| models | Cross0 | Cross1 | Cross2 | Cross3 | Cross4 | Mean |
|---|---|---|---|---|---|---|
| neuron dropout (%) | 54.40 | 40.28 | 52.38 | 41.88 | 51.56 | 48.10 |
| grid dropout (%) | 60.29 | 42.08 | 56.42 | 48.97 | 50.91 | 51.73 |

### 4.4. Neuron Dropout and Grid Dropout without Fine-Tuning

In order to analyze the relationship between neuron dropout and grid dropout, we carry out experiments without finetuning the fully connected layers from existing models trained on ImageNet. But instead, we just fine-tune all the convolutional layers for the VGG-net, and train the fully connected layer from scratch. Similar to the comparison in Section 4.3, we give the comparison on accuracy, training and testing loss curves, and gradCAM. The results are shown in Table 2, Fig. 4, and Fig. 6. Compared with neuron dropout, grid dropout achieves better classification result, strong generalization ability, and better gradCAM. We believe that, the inferior performance of neuron dropout is attributed to insufficient training data which can hardly train a more complex ensemble classifier. From Fig. 6, we observed similar patterns as in Section 4.3. Interestingly, we notice that neuron dropout (second row) is more likely to overfit to certain areas of human face, e.g. the forehead of examples 4-7. In comparison, our grid dropout (third row) successfully captures the whole face for almost all examples even under insufficient training samples.

### 4.5. Comparison with State-of-the-Arts

The third group of experiments are conducted to demonstrate our model's superiority to the state-of-the-art methods. In this section, since our proposed methods use plain CNN model without any bells and whistles, we merge ordinal information to the group classification with an additional regression loss, i.e., Eq. 2. From Table 3, our final model beats all competitors with exception on [3]. We believe the advantage of [3] is gained from pre-training on a large facial dataset MS-Celeb-1M [2]. It is very likely our model will benefit from the same advantage by doing the same pre-training.

**Table 3**. A comparison with existing methods.

| models | Mean |
|---|---|
| LBP+FPLBP [15] (%) | 45.10 |
| Deep CNN [1] (%) | 50.70 |
| Cumulative Attribute [2] + CNN (%) | 52.34 |
| L-w/o-hyper [3](pretrained on MS-Celeb-1M) (%) | 49.46 |
| L-w/o-KL [3](pretrained on MS-Celeb-1M) (%) | 54.52 |
| Full model [3](pretrained on MS-Celeb-1M) (%) | 56.01 |
| Our proposed model (%) | 54.20 |

### 5. CONCLUSION

In this paper, we propose a method of grid dropout that randomly drops out some patches of raw images. In order to keep the image's structure, masking label is used as target for training simultaneously. We show our proposed methods from three aspects: accuracy performance, generalization ability, and visualization of gradCAM. We also discuss the relationship between neuron dropout and grid dropout and conclude that, for small or medium dataset, grid dropout is better than neuron dropout due to less parameters for learning. Finally we compare our model with state-of-the-art methods and witness very competitive results.

---

[1]For the sake of saving memory, tensorboard automatically sampled 1000 records of loss and accuracy out of 30,000 training iterations with uniform interval, i.e. we ultimately trained our model for 30,000 iterations.

[2]MS-Celeb-1M is large enough, the pre-training experiment on all the methods need more than half month. Our result has not been obtained on time for Camera-Ready version. It will appear at the extended journal version.

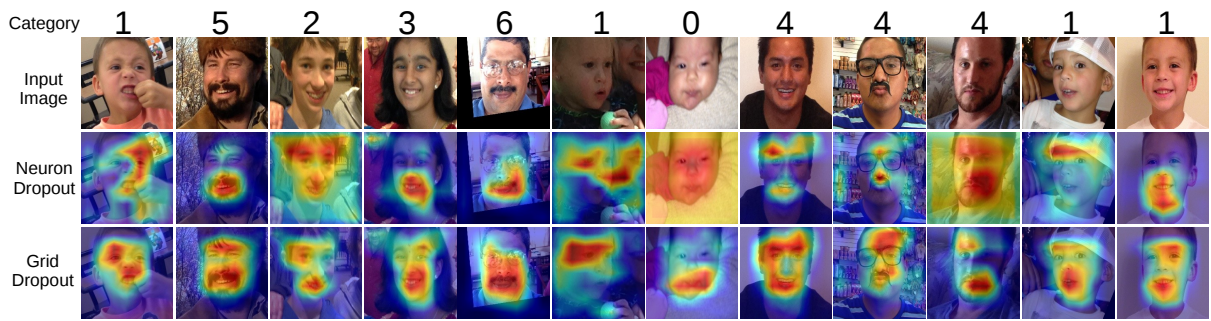

**Fig. 5**. The gradCAM comparison between neuron dropout and grid dropout. Three rows respectively represent input image, gradCAM of neuron dropout and gradCAM of grid dropout. Each image's ordinal category label is attached to the top of input image.(Best viewed in color.)

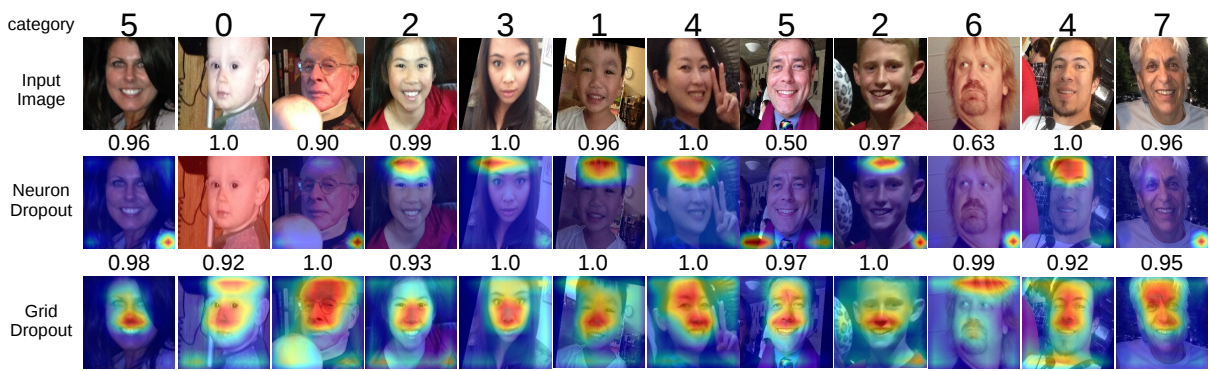

**Fig. 6**. The gradCAM comparison between neuron dropout and grid dropout without fine-tuning the fully connected layers. Three rows respectively represent each input image, gradCAM of neuron dropout and gradCAM of grid dropout. The maximum class probability is attached to the top of both neuron and grid dropout gradCAM images. (Best viewed in color.)